\documentclass[journal]{vgtc}                     

\onlineid{1603}

\vgtccategory{Research}
\vgtcpapertype{Applications}

\newcommand{\tool}{\emph{ModalChorus}\xspace}

\title{\tool: Visual Probing and Alignment of Multi-modal Embeddings via Modal Fusion Map}

\author{%
  Yilin Ye,
  Shishi Xiao,
  Xingchen Zeng, and
  Wei Zeng
}

\authorfooter{
  \item
  Y.Ye, S. Xiao, X. Zeng, and W. Zeng are with the Hong Kong University of Science and Technology (Guangzhou) . E-mail: \{yyebd@connect., sxiao713 @connect., xzeng159@connect., weizeng@\}hkust-gz.edu.cn.
  Y. Ye and W. Zeng are also with the Hong Kong University of Science and Technology.
  \item
    Wei Zeng is the corresponding author.

}

\abstract{
Multi-modal embeddings form the foundation for vision-language models, such as CLIP embeddings, the most widely used text-image embeddings.
However, these embeddings are vulnerable to subtle misalignment of cross-modal features, resulting in decreased model performance and diminished generalization.
To address this problem, we design \tool, an interactive system for visual probing and alignment of multi-modal embeddings.
\tool primarily offers a two-stage process:
1) embedding probing with Modal Fusion Map (MFM), a novel parametric dimensionality reduction method that integrates both metric and nonmetric objectives to enhance modality fusion;
and 2) embedding alignment that allows users to interactively articulate intentions for both point-set and set-set alignments.
Quantitative and qualitative comparisons for CLIP embeddings with existing dimensionality reduction (\eg, t-SNE and MDS) and data fusion (\eg, data context map) methods demonstrate the advantages of \emph{MFM} in showcasing cross-modal features over common vision-language datasets.
Case studies reveal that \tool can facilitate intuitive discovery of misalignment and efficient re-alignment in scenarios ranging from zero-shot classification to cross-modal retrieval and generation.
}


\teaser{
  \centering
  \includegraphics[width=\linewidth]{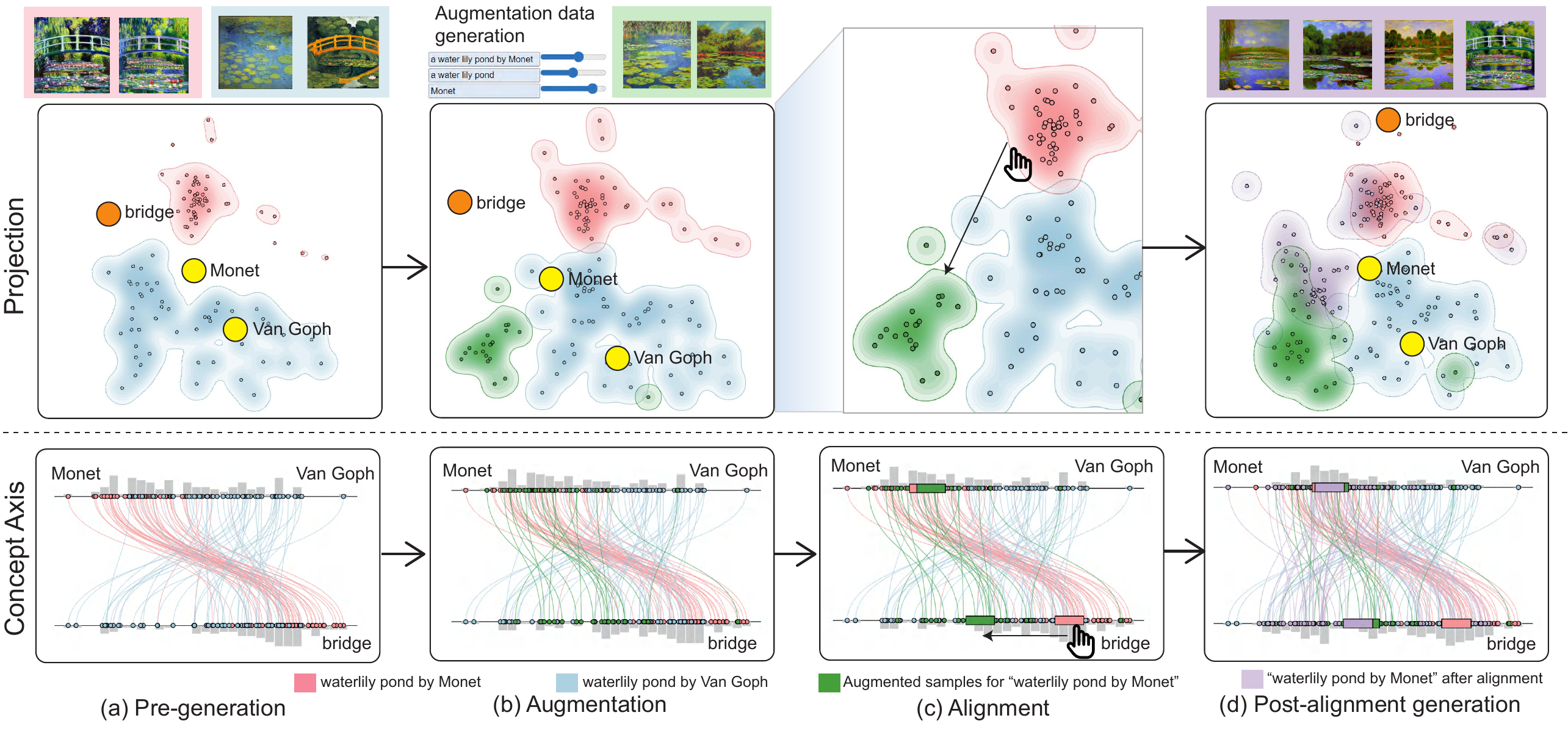}
  \vspace{-6mm}
  \caption{
  Visual probing of multi-modal CLIP embeddings for text-to-image generation. (a) For the prompt of "waterlily pond by Monet", users first discover misalignment of pre-trained models in the form of concept entanglement between \emph{"Monet"} and \emph{"bridge"} using our Modal Fusion Map projection and concept axis view. (b) Data augmentation based on weighted embedding generation can be performed to provide extra alignment reference set. (c) Set-set alignment interaction is performed to align the initial generated images of "waterlily pond by Monet" to the augmented images reflecting user intents. (d) Post-alignment model can generate a set of images with more diversity by disentangling \emph{"Monet"} and \emph{"bridge"}.
  }
  \label{fig:teaser}
}

\graphicspath{{figs/}{figures/}{pictures/}{images/}{./}} 

\usepackage{tabu}                      
\usepackage{booktabs}                  
\usepackage{lipsum}                    
\usepackage{mwe}                       
\usepackage{amsmath}
\usepackage{amssymb}
\usepackage{multirow}
\usepackage{mathptmx}  

\newcommand{\eg}{\emph{e.g.}}

\newcommand{\etal}{\emph{et al.}}

\usepackage{colortbl}
\definecolor{mred}{rgb}{.80,.12,.30}
\definecolor{grey}{rgb}{0.5,0.5,0.5}
\definecolor{lgrey}{rgb}{0.7,0.7,0.7}
\definecolor{purple}{rgb}{.75,0,.85}
\definecolor{pistachio}{rgb}{0.58, 0.77, 0.45}
\definecolor{myorange}{rgb}{0.94, 0.36, 0.13}

\begin{document}



\maketitle


\section{Introduction}
Neural embeddings are high-dimensional latent representations for knowledge captured from self-supervised pre-training, such as word embeddings and image embeddings.
Recently, multi-modal (\eg, text and image) embedding are playing a pivotal role for advancing multi-modal AI models.
This type of embeddings learns a joint representation space that encodes different modalities and their relationships, forming the basis for cross-modal tasks such as text-to-image retrieval and generation~\cite{baldrati2022conditioned, ramesh2022hierarchical, ye2023contemporary, ye2024generative}.
The performance of multi-modal embedding models rely heavily on the quality of multi-modal alignment, which seeks to match data with corresponding semantics across different modalities within the embedding space~\cite{radford2021learning, gao2023softclip, girdhar2023imagebind}.
However, misalignment in multi-modal embeddings is common due to the intricate many-to-many mapping among concepts in different modalities.
For instance, text-to-image embeddings can easily encounter misalignment issues of concept entanglement.
As illustrated in Figure~\ref{fig:teaser}(a), the text prompt of `\emph{waterlily pond by Monet}' becomes entangled with the `\emph{bridge}' concept in the image modality, reducing the diversity of generated images.

Identifying misalignment in multi-modal embeddings is crucial for enhancing model performance.
Existing methods for evaluating misalignment often rely on reference-based evaluations (\eg, CIDEr~\cite{vedantam2015cider} and SPICE~\cite{anderson2016spice}) that necessitate extensive human-labeled references, or reference-free metrics (\eg, CLIPScore~\cite{hessel2021clipscore}) derived from pretained multi-modal models.
Despite not relying on references, reference-free metrics are reliant on pretrained models, making it challenging for fully automatic methods to detect misalignment in diverse and context-dependent scenarios. 
For instance, the CLIPScore for the text prompt `\emph{waterlily pond by Monet}' fails to reflect the issue of concept entanglement, as the CLIP model itself is biased towards the `\emph{bridge}' concept in the image modality.
Hence, existing fine-tuning techniques to improve alignment often fall short of expectations in numerous scenarios. 
There is a need for an interactive visualization tool to help users intuitively investigate and address misalignment.

However, the intricate data structures and feature characteristics inherent in multi-modal embeddings pose particular challenges for visual probing and interactive alignment.
A key challenge arises from the modality gap, wherein embedding vectors from different modalities are essentially disjointed in the joint embedding space~\cite{liang2022mind}.
To achieve cohesive visualization of multi-modal embeddings, it is essential to address the modality gap issue and unify the presentation of different modalities within a single display space.
Previous visualizations of neural embeddings have primarily centered on single-modal embeddings, such as word or image embeddings~\cite{liu2017visual, liu2019latent, heimerl2018interactive}.
Notably, these works commonly employ classical dimensionality reduction (DR) methods like t-SNE~\cite{van2008visualizing} and MDS\cite{borg2005modern}, which are limited to separately displaying multi-modal embeddings in distinct spaces.
Fusion-based DR methods (\eg,~\cite{choo2012heterogeneous, cheng2015data}) offer a potential solution to jointly project embeddings from different modalities.
However, these methods typically treat intra- and inter-modal distances equally, without giving special consideration to cross-modal relations.
For example, Data Context Map (DCM)~\cite{cheng2015data} solely relies on metric-based objectives that poorly capture the relative rank order of inter-modal distances.
As illustrated in Figure~\ref{fig:projection_compare2}(left), DCM projects rather even distribution of image embedding points around the textual concepts, making it harder to observe differences in distribution pattern.

Moreover, enabling interactive alignment for multi-modal embeddings presents another challenge, primarily due to two reasons.
First, user-intended alignment strategies encompass diverse operations.
For example, in Figure~\ref{fig:teaser}, upon identifying the concept entanglement between `\emph{Monet}' in the text modality and `\emph{bridge}' in the image modality, users may prefer to drag the `\emph{bridge}' point far away or relocate the entire set of `\emph{Monet}' images.
However, existing studies often focus on point-based operations~\cite{endert2011observation, xia2022interactive}, while others solely support set-based interaction~\cite{fujiwara2021interactive}.
Secondly, users would utilize interactive alignment to refine the underlying models and ensure that the refined model performs as expected, as illustrated by the disentangled images generated post-alignment, as in Figure~\ref{fig:projection_compare2}(d).
Existing studies on DR refinement mostly focus on adapting the projection layout~\cite{endert2011observation, xia2022interactive, wang2023drava, fujiwara2021interactive}, whilst overlooking the model refinement.

To fill the gap, we present \tool, an interactive system that supports visual probing and alignment of multi-modal embeddings.
\tool mainly comprises two-stage exploration.
First, in the \emph{embedding probing} stage, we propose Modal Fusion Map (MFM), a novel parametric DR method integrating metric and non-metric objectives for enhanced modality fusion.
By taking the advantages of metric-based objectives in preserving the intra-modal distances and non-metric-based objectives in capturing inter-modal distance rank order~\cite{borg2005modern, choo2012heterogeneous}, MFM effectively addresses the modality gap challenge induced by multi-modal embeddings.
Compared with conventional single-modal and fusion-based DR methods, MFM achieves higher trustworthiness and continuity regarding inter-modal relations (see Table~\ref{tab:proj_tab}), and can better visually reflect the intra- and inter-modality contextual distributions (see Figures~\ref{fig:projection_compare1} \& \ref{fig:projection_compare2}).
Next, in the \emph{embedding alignment} stage, to accommodate the diverse alignment scenarios, we design an alignment interaction scheme that allows for alignment on multiple levels including point, subset, and set.
The interaction scheme is integrated with MFM encompassing point-set and set-set alignment.
Besides, a concept axis view is also developed to enable linear visual representation for the probing and alignment of multi-modal embeddings.

In summary, our make the following contributions:

\begin{itemize}
    \item
    We propose Modal Fusion Map (MFM), a novel dimensionality reduction method tailored for fusion projection of multi-modal embeddings.
    The effectiveness of MFM is demonstrated using both quantitative and qualitative evaluations.
    
    \item 
    We develop \tool, an interactive system that supports visual probing of multi-modal embeddings to discover misalignment, along with an interaction scheme that supports interactive fine-tuning of the underlying multi-modal embedding models.

    \item 
    We show the effectiveness of our system through case studies on three embedding-based cross-modal tasks, ranging from zero-shot classification to cross-modal retrieval and generation.

\end{itemize}
\section{Related Work}

\begin{figure*}[t]
\centering
\includegraphics[width=0.99\textwidth]{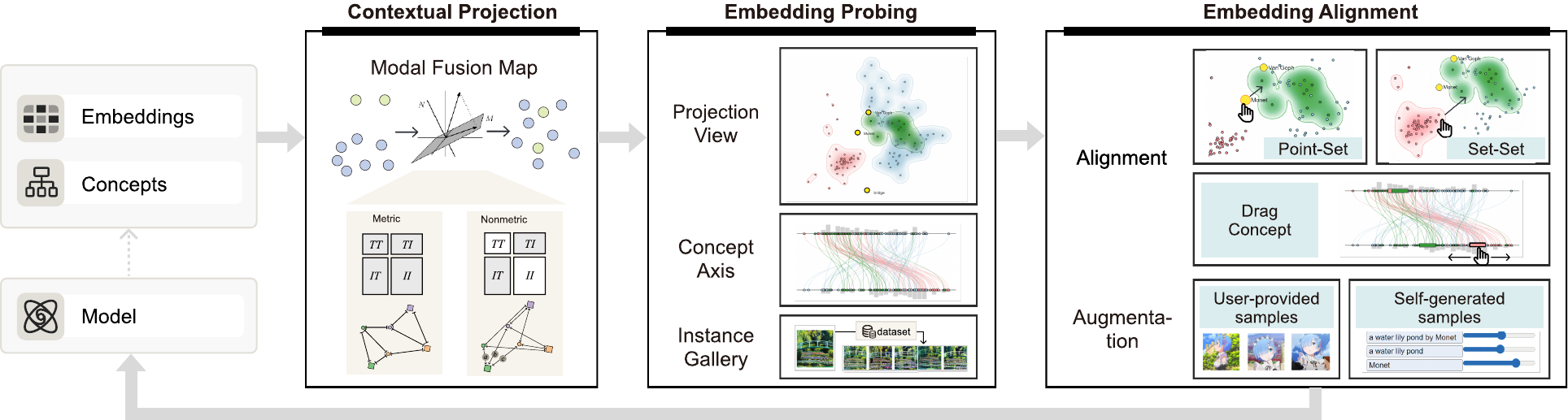}
\vspace{-2mm}
\caption{
Overview of our framework.
Multi-modal embeddings and concepts extracted from text and images are first projected with Modal Fusion Map, a novel modality-fusing DR method we propose.
In the visual exploration stage, visual probing of the embeddings is enabled in the projection view and the concept axis view, allowing users to explore embedding sets and individual instance point.
Finally, in the embedding alignment stage, interactive alignment with point-set and set-set alignment schemes is provided, along with optional augmentation with few-shot samples.
}
\label{fig:overview}
\vspace{-3mm}
\end{figure*}

\noindent
\textbf{Visualization for Neural Embeddings}.
Deep learning relies on neural networks that are often pre-trained on large amounts of data.
Neural embeddings are the foundational high-dimensional feature representation of raw data encoded by neural networks, such as text embeddings like word2vec~\cite{mikolov2013distributed} and BERT~\cite{devlin2018bert} and image embeddings like SimCLR~\cite{chen2020simple}.
Visualization researchers have dedicated significant efforts to enhance the comprehension of neural embeddings.
Previous studies primarily focus on unimodal embeddings, encompassing word embeddings~\cite{liu2017visual, heimerl2018interactive, heimerl2020embcomp} and image embeddings~\cite{liu2019latent}.
Many of these studies integrate projection methods with axis-based~\cite{liu2017visual, heimerl2018interactive, liu2019latent} or set-based~\cite{heimerl2020embcomp} exploration techniques.
For example, Liu \etal~\cite{liu2017visual} identified analogy axis between multiple pairs of words with the same semantic transition in word embeddings projected by t-SNE.
Latent Space Cartography~\cite{liu2019latent} extends the concept of semantic axis to customized axis defined by users, which can be applied to exploration of both unimodal word embeddings and image embeddings.
EmbComp~\cite{heimerl2020embcomp} combines t-SNE projection with visualization of neighborhood set overlap to compare different word embedding models.

Recently, multi-modal embeddings such as CLIP~\cite{radford2021learning} and ALIGN~\cite{jia2021scaling} have fueled the advances in multi-modal AI such as text-to-image generation. 
These embeddings can encode data from different modalities in a joint space, contributing to various applications such as cross-modal retrieval~\cite{baldrati2022effective} and generation~\cite{ramesh2022hierarchical, zeng2024advancing}.
However, this integration also introduces modality gap~\cite{liang2022mind} that signifies discrepancies between different modal embeddings, complicating the comprehension of multi-modal embeddings.
There is a lack of visualization tool tailored to the task.
Specifically, the task demands an effective visualization method for probing multi-modal embeddings and an interactive scheme for improving alignment of multi-modal embeddings. 
To meet the goal, we propose the Modal Fusion Map that can better preseve the contextual information of multi-modal embeddings, and an interactive alignment scheme that offers visual steering for modal alignment.

\smallskip
\noindent
\textbf{Contextual Dimensionality Reduction}.
Dimensionality reduction for multi-modal data has been a challenging problem as traditional DR methods like t-SNE~\cite{van2008visualizing, skrodzki2024accelerating}, PCA~\cite{wold1987principal}, and MDS~\cite{borg2005modern} cannot account for cross-modal relations due to the modality gap~\cite{liang2022mind, zhou2023clip, ouali2023black}.
Contextual visualization is a type of DR method designed to project data points in relation to attribute points~\cite{cheng2015data, zhang2022graphical, meyer2002generalized}, which can be applied to multi-modal data projection, yielding more integrated visualization than dual analysis~\cite{dennig2023fs}.
Existing contextual visualizations can be categorized into two types: anchor-based projection and fusion-based methods.
Anchor-based methods employ a two-stage approach, initially determining the layout of points in one modality before calculating the position of points in the other modality.
For example, the RadViz method~\cite{hoffman1997dna, cheng2017radviz, ye2022visatlas} first lays out the attribute points on a circle and then projects the data points based on their multi-dimensional attribute values.
However, the structure of the embedding space can be significantly distorted due to the challenge of optimally laying out the anchor points.

One type of fusion methods, known as co-embedding methods~\cite{xie2018semantic, choi2023co}, introduces their own high-dimensional representations of multi-modal data or modifies the embeddings of certain data points to achieve a desired visual layout.
However, these methods diverge from our goal as they alter the original embeddings with custom models, which cannot help users understand commonly used multi-modal embeddings in AI tasks.
Other fusion methods are limited to more specific conditions~\cite{globerson2004euclidean, wu2017computation, gonen2014embedding}, such as COPE~\cite{globerson2004euclidean} which requires co-occurrence statistics.
Visualization researchers have developed more general fusion methods~\cite{cheng2015data, zhang2022graphical}. 
Particularly, Data Context Map (DCM)~\cite{cheng2015data} defines the distance matrix for the attributes and merges it with the data distance matrix before using MDS to jointly project the attribute points and data points.
However, in DCM, intra-modal and inter-modal distances are equally treated in metric-based optimization, which limits its ability to adequately capture the cross-modal non-metric ordinal structure of multi-modal embeddings.
In our study, we introduce the Modal Fusion Map, which integrates both metric and nonmetric objectives into modality fusion using a novel parametric DR method, to effectively preserve relationships for intra- and inter-modal distances of multi-modal embeddings.

\smallskip
\noindent
\textbf{Visual Steering for Modal Alignment}.
Pre-training of multi-modal foundation embedding models relies on alignment through methods like contrastive learning, such as ViLBERT~\cite{lu2019vilbert} and CLIP~\cite{radford2021learning}.
For example, the CLIP embeddings~\cite{radford2021learning} is pre-trained on large-scale image-text pair corpus by matching image and text caption in a joint representation space with contrastive multi-class N-pair loss.
The pre-training methods typically aim to establish a foundational model,
yet the varying quality of pre-training data often leaves some misalignment in specific cases, which requires adaptation such as few-shot fine-tuning~\cite{hu2021lora, gal2022image, ouali2023black} to refine the alignment.
Misalignment cases may require human knowledge to be discovered, and the fine-tuning process also typically involves users' choice of alignment data and direction, which necessitates an interactive system to support human-in-the-loop workflow.
This scenario differs from interactive prompt engineering of pre-trained models~\cite{wu2022promptchainer, strobelt2022interactive, feng2023promptmagician, guo2024prompthis}, as prompt engineering only seeks to alter the input without refining the model, which is not enough for steering complex multi-modal models with misalignment.

Some previous visual analytics systems support interactive improvement of AI models through label correction or data augmentation~\cite{chen2021towards, gou2020vatld, he2021can}.
For example, VATLD~\cite{gou2020vatld} leverages disentangled representation learning for semantic exploration of traffic light detection results in relation to explainable data dimensions.
However, these studies only focus on task-specific models without paying attention to foundational embeddings~\cite{yang2024foundation}.
Many studies also rely on the ground-truth labels for insight discovery, which may not be available in real-time probing of pre-trained models.
In addition, these studies lack support for visual steering interaction directly in the visualization space, which is more intuitive for the alignment operation our study aims at.

Some visualization researchers have studied interactive visual steering of dimensionality reduction results~\cite{endert2011observation, xia2022interactive, wang2023drava, fujiwara2021interactive}.
For example, Xia \etal~\cite{xia2022interactive} proposed a contrastive learning-powered parametric dimension reduction method to support point-level interaction to enhance the visual clustering effect.
ULCA~\cite{fujiwara2021interactive} supports set-level visual steering interaction for comparative analysis.
DRAVA~\cite{wang2023drava} introduces an interaction method to adjust the positions of small multiples in axis-based visualization based on $\beta VAE$.
However, these interactions only focus on refining the projection layout for visual exploration purposes, lacking the ability to align the underlying models or high-dimensional representations.
In addition, the interaction schemes of most previous studies are limited to a single type of interaction, such as point-based or set-based interaction in a single view, which cannot cover the diverse alignment scenarios of multi-modal embeddings.
In our study, we develop an interaction scheme supporting point-set and set-set alignments, enabling flexible alignment of underlying embedding-based models.

\section{Overview}
\begin{figure*}[t]
\centering
\includegraphics[width=0.998\textwidth]{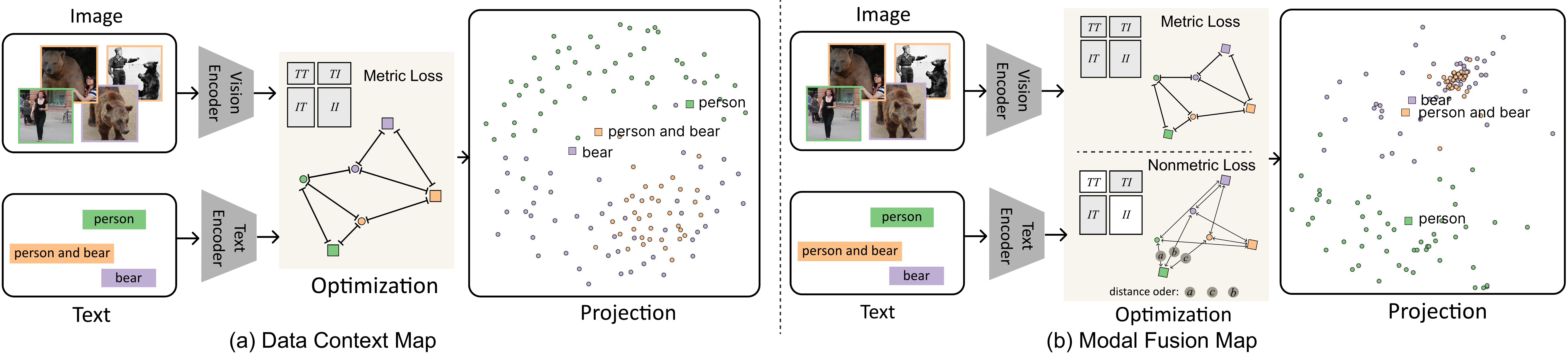}
\vspace{-6mm}
\caption{
(a) Data Context Map (DCM) only considers metric-based optimization that indiscriminately seeks to preserve the absolute distance for intra-modal and inter-modal pairs of data points. (b) Inspired by the observation that the nonmetric rank order of cross-modal distances is important for multi-modal embedding-based tasks, our Modal Fusion Map (MFM) combines metric and nonmetric objectives for fusion.
}
\label{fig:compare_method}
\vspace{-4mm}
\end{figure*}

\subsection{Background and Domain Problem}\label{ssec:task}
\noindent
\textbf{Multi-modal embedding}.
Multi-modal embedding models are pre-trained encoder models for the representation of multi-modal data.
For example, the CLIP model is pre-trained on a large corpus of image-text pairs, using transformers and vision transformers to first separately encode text and image into high-dimensional vectors.
Then, through a linear transformation, the text embedding and image embedding vectors are aligned in a shared embedding space with contrastive loss.
Multi-modal embeddings are the foundational encoder for many AI tasks that involve multi-modal data in its input and/or output. 
Common tasks include semantics-based image classification~\cite{radford2021learning}, cross-modal retrieval~\cite{girdhar2023imagebind}, and text-to-image generation~\cite{rombach2022high, chen2023vlp}.

\smallskip
\noindent
\textbf{Alignment}.
In multi-modal models, alignment means the matching of data representations with corresponding semantics from different modalities.
In the pre-training stage of CLIP, for example, the alignment is achieved by updating the embeddings of an image and its corresponding text caption so that they are closer than incorrect pairs in the high-dimensional representation space.
However, due to the varying quality and large quantity of data in pre-training and imperfection in training algorithms, there may be misalignment in the pretrained model, which requires further adaptation for enhancing alignment~\cite{hu2021lora, gal2022image, ouali2023black}. 

\smallskip
\noindent
\textbf{Multi-level Alignment}.
There is mainly alignment on three levels: point, subset, and set, requiring two types of alignment: point-set alignment and set-set alignment.
Specifically, users may discover misalignment of an individual point (\eg, misclassified image point or misunderstood text point), a subset (\eg, a subset of incorrect samples in the whole set of text-to-image retrieval results), and a set (\eg, biased or entangled generation results of text-to-image models), requiring different alignment operations.
To clarify, we refer to keywords extracted by our system or entered by users as concepts, which are the main text embeddings we focus on in this study for contextual exploration of embeddings, while particular image point is referred to as instances.

\smallskip
\noindent
\textbf{Problem: visualization for embedding}.
Many visualization studies treat embedding methods as a tool for processing data, with the aim of optimizing the visual display of raw input data.
That is, these visualizations regard embedding as a projection method.
Instead, in the field of AI, representation learning of single and multi-modal embeddings has been playing a pivotal role for various downstream tasks~\cite{ruder2019survey, le2020contrastive}. 
The high-dimensional embeddings themselves are the key intermediate representations of data extracted from raw text or pixels, not just the representation for visual display only.
To gain insight into large AI models, particularly for the alignment problem, high-dimensional visualization methods should \emph{prioritize capturing the features of the foundational embeddings itself (\textbf{G1})}.
For example, suppose two classes of images are indeed close in the embedding space, which signifies risks of misalignment in the embedding. In that case, we do not wish to maximize the class separation in projection space just for visual display since it can mislead users.
For another example, if the prompt's text embedding in the generation model is not close to the generated images, we should be cautious about directly putting the text at the centroid of the generated image set, which may lead users to believe that the generation is fully aligned.  

To summarize, the former studies focus on embedding for visualization, while our work aims at visualization for embedding.
In addition, the aim for interaction in this scenario is to \emph{improve the foundational high-dimensional embeddings (\textbf{G2})} instead of improving the visual display of data like previous studies did~\cite{xia2022interactive, fujiwara2021interactive}.

\subsection{Challenges and Design Requirement}

Accomplishing these two goals is challenging for existing visualization methods, particularly due to:

\begin{enumerate}[leftmargin=*, label=\textbf{C\arabic*}]
\vspace{-0.5mm}
\item
\textbf{Modality Gap}.
The heterogeneous distributions of different modalities in the joint embedding space result in the modality gap~\cite{liang2022mind, zhou2023clip}, making it difficult for existing DR methods to simultaneously capture intra-modal and cross-modal features.

\vspace{-0.5mm}
\item
\textbf{Diverse alignment intentions}.
The diverse alignment scenarios in different cross-modal tasks post challenges to designing a comprehensive interaction scheme integrated into the visualization.

\end{enumerate}

To tackle the challenges, we summarize the design requirements of \tool, 
which should support flexible and effective \emph{R1) visual probing} of multi-modal embeddings to meet \emph{\textbf{G1}}.

\begin{enumerate}[leftmargin=*, label=\textbf{R1.\arabic*}]
\vspace{-0.5mm}
    \item \textbf{Accurately preserving inter- and intra-modal distances.} An effective fusion-based DR method is needed to bridge the modality gap while maximally preserving inter- and intra-modal relations.    

\vspace{-0.5mm}
    \item \textbf{Effective visual presentation to help identify misalignment.} Apart from the projection, effective graphical enhancement is needed to assist discovery of misalignment issues such as misclassification or entanglement. 
\end{enumerate}

Second, \tool shall facilitate \emph{R2) interactive alignment} of multi-modal embeddings to support \emph{\textbf{G2}}:
\begin{enumerate}[leftmargin=*, label=\textbf{R2.\arabic*}]
\vspace{-0.5mm}
    \item \textbf{Supporting alignment on point and set levels.} Users may discover embedding misalignment on an individual data point or a whole set of points, demanding different types of alignment interaction, including point-set and set-set alignment.

\vspace{-0.5mm}
    \item \textbf{Supporting axis-based alignment.} 
    Previous embedding visualization studies have identified the semantic axis as an effective complement of the overall projection for more focused concept-related exploration~\cite{liu2017visual, heimerl2018interactive, liu2019latent}.
    Besides directly manipulating the projection of embeddings, users also need to perform axis-based alignment as the axis can more clearly show the direction of alignment with respect to a specific semantic concept. 

\vspace{-0.5mm}
    \item \textbf{Supporting data augmentation.} When users discover misalignment but cannot find correct reference data, they would like to provide extra data and process it to help the alignment.
\end{enumerate}

\subsection{ModalChorus Overview}

An overview of our system is shown in Fig.~\ref{fig:overview}, which mainly consists of two stages: 1) embedding probing and 2) embedding alignment.
In the first stage, starting from a particular dataset and task, along with user-provided input or automatically extracted concept, we support visual probing of the embeddings with sampled data for interpretation of embeddings and discovery of misalignment.
Particularly, we develop Modal Fusion Map, a novel parametric fusion method that integrates metric and nonmetric objectives for multi-modal embedding projection.
We also incorporate a concept axis view that allows users to explore the correlation of image embeddings in relation to concept text embeddings.
An additional instance gallery displays similar images to the selected image point in the embedding space for neighborhood exploration.

In the second stage, upon discovering misalignment, we enable users to select a particular point, subset, or set and perform point-set alignment or set-set alignment in either the projection view or the concept axis view.
In some cases, when new data is needed to enhance the alignment, we allow users to upload their collected data for few-shot alignment or use our system's weighted embedding generation function to generate candidate augmentation data.
Finally, the visual alignment operations are mapped to the backend fine-tuning.

\section{Multi-modal Contextual Visualization}

In this section, we describe Modal Fusion Map, a novel DR method we propose to address \emph{R1 visual probing} of multi-modal embedding.

\subsection{Problem Identification}\label{ssec:problem}

\begin{figure*}[t]
\centering
\includegraphics[width=0.995\textwidth]{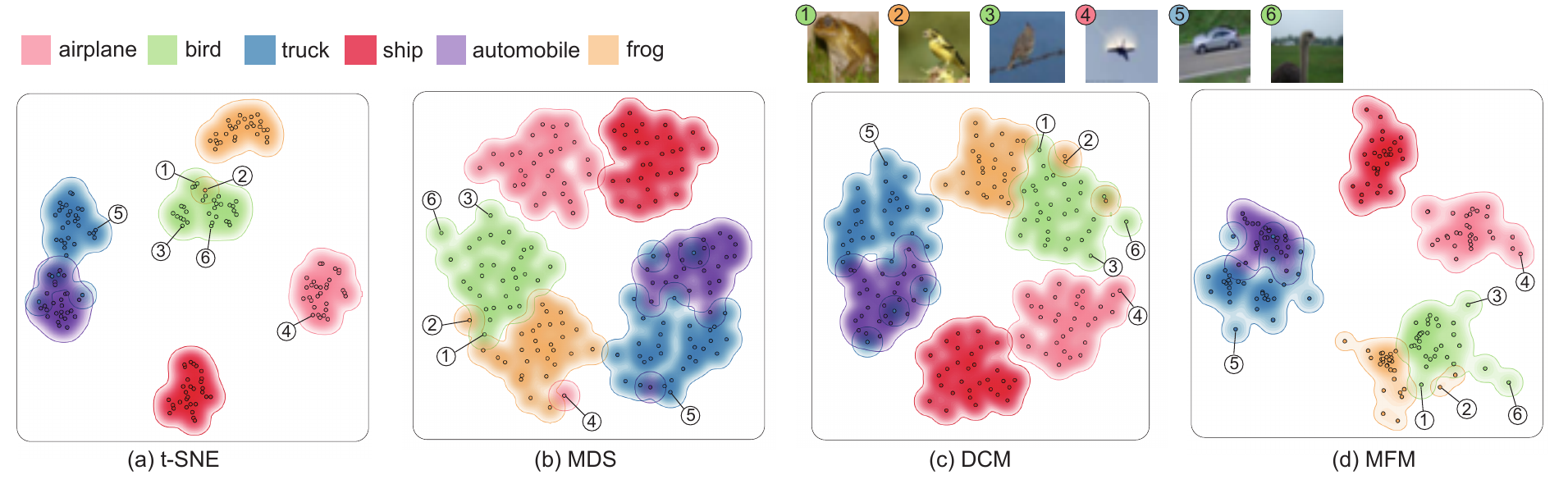}
\vspace{-4mm}
\caption{
For the zero-shot classification task on CIFAR-10 which relies on cross-modal similarity (color of points represent the predicted class), MFM can better reflect set relations and outliers for visual probing of misalignment.
}
\label{fig:projection_compare1}
\vspace{-4mm}
\end{figure*}

To address the modality gap problem in multi-modal embedding visualization, data matrix fusion methods~\cite{choo2012heterogeneous, cheng2015data} are a promising solution.
Matrix fusion methods such as Data Context Map (DCM)~\cite{cheng2015data} are derived from the MDS method for distance-based fusion. 
The original Data Context Map is designed for the attribute and data spaces of multi-dimensional data.
Specifically, to align data points from different modalities, it constructs a large distance matrix containing the pairwise distances between all the data points and attribute points, where the intra-modal distance is the original high-dimensional distance such as Euclidean or Cosine distance while the cross-modal distance needs to be defined according to data properties.
For example, DCM defines the distance between attribute point and data point as $1-v$, where $v$ is the data point's value in this attribute dimension.

First, to account for high dimensional latent space, we can naturally change the attribute-data distance in DCM to the Cosine distance between text embedding and image embedding.
However, this modification may not suffice for the complexity of multi-modal embedding.
Specifically, to enhance the modality merging effect, it is important to flexibly adjust the weights of intra-modality and inter-modality distance. 
Directly scaling the submatrix as mentioned in\cite{choo2012heterogeneous} may have the risk of significantly distorting the embedding space or exacerbating the modality gap.
More importantly, when multi-modal embeddings like CLIP are used for cross-modal tasks such as text-to-image retrieval, the absolute distance between the text and image embeddings is less important than the relative order of the distance values.
In visualization of multi-modal embeddings, such characteristics should also be considered.
This means that we should develop a better fusion method that considers both metric and non-metric objectives~\cite{hout2013multidimensional, quist2004distributional, faith1987compositional}.

\subsection{Modal Fusion Map}\label{ssec:multi_project}

\noindent
\textbf{Dimensionality Reduction}.
Inspired by recent work on parametric dimensionality reduction~\cite{xia2022interactive, xia2023parallel, zang2022evnet}, to satisfy the projection requirements presented above, we propose Modal Fusion Map (MFM) which can flexibly combine different objectives for joint multi-modal embedding projection.
The hypothesis is that in the high dimensional embedding space, there exists a subspace or manifold surface $S$ between the text embeddings set $T$ and image embeddings set $I$, such that the projection of embeddings from both modalities on this surface ($P(T), P(I) \in S$) can result in an optimized 2D parametric representation $S(x, y)$.
Specifically, like other matrix-based methods, we first compute the merged distance matrix: 
\begin{gather}
M=
\begin{pmatrix}
II, & IT \\
TI, & TT
\end{pmatrix},
\end{gather}
where $II$ is image distance submatrix, $TT$ is text distance submatrix, $IT=TI^T$ is the cross-modal distance submatrix, all using cosine distance or the equivalent Euclidean distance between normalized vectors.
Each submatrix is normalized by their mean value.

Next, instead of directly applying the traditional MDS method to the merged matrix as in DCM, we parametrize the projection with a three-layer feed-forward neural network mapping 512 or 1024-dimensional CLIP embedding to the 2-dimensional projection space.
See supplementary material for more detail.
Then, to implement the MDS objective, we construct a loss function using the Pearson correlation between the high dimensional merged distance matrix and the projected distance matrix for scale-free optimization.
\begin{gather}
L_{M}=-\frac{\sum(M_{i,j}-\overline{M})(P_{i,j}-\overline{P})}{\sqrt{\sum(M_{i,j}-\overline{M})^2}\sqrt{\sum(P_{i,j}-\overline{P})^2}},
\end{gather}
where $P$ stands for the distance matrix of the projected points.

In this way, we can easily define loss terms for the intra-modality and inter-modality submatrices, denoted as $L_{TT}$, $L_{II}$, and $L_{IT}$, respectively.
Accordingly, the loss function for metric MDS is the weighted sum. 
In our case, we only consider the overall term and the cross-modal term: $L_1=w_{1}L_M+w_{2}L_{IT},$
where we set $w_{1}=10, w_{2}=2$.

In addition, for the nonmetric loss to preserve cross-modality distance order, we further introduce another loss term:
\begin{gather}
L_2=\frac{-\sum_{j<k} f((TI_{i,j}-TI_{i,k})*(P(TI)_{i,j}-P(TI)_{i,k}))}{\|P(TI)\|_{2}},
\end{gather}
where $f(x)= \begin{array}{l}
0, x\ge0 \\
-x, x<0
\end{array}$.
This loss term will be zero when all the cross-modal distance order is preserved in the projection.
The final loss $L=L_{1}+\alpha L_{2}, \alpha=0.05$.
$w_1$, $w_2$, $\alpha$ are selected empirically. 
Code is available at: \url{https://github.com/yilinye/Modal-Fusion-Map}.

\smallskip
\noindent
\textbf{Contour-based graphical enhancement}. 
We provide graphical enhancements in the form of density contour as inspired by recent work~\cite{zhang2022graphical}. 
As shown in Fig.~\ref{fig:projection_compare2}, the density plot can show the default KDE density estimation of data point distribution.
The KDE contour can serve as a graphical representation of sets in the projection view, which can facilitate subsequent alignment interaction as we describe below.
Alternatively, when users provide customized metrics defined for the data points, such as CLIP-Score for generated samples, the density plot can show the kernel estimation of the metric value distribution.

\begin{table}
\scriptsize
\centering
\caption{Evaluation of projection methods with inter-modal and intra-modal trustworthiness (T) and continuity (C) metrics.}
\begin{tabular}{c|cc|cc}
\hline
& \multicolumn{2}{c|}{\textbf{Inter-modal}} &  \multicolumn{2}{c}{\textbf{Intra-modal}} \\
       & T(30) & C(30)  & T(30) & C(30) \\\hline
PCA &  0.9177 & 0.9301  & 0.7297 & 0.8183 \\
MDS  & 0.9274 & 0.9336  & 0.8039 & 0.8537 \\
Isomap   & 0.9307 & 0.9281  & 0.7706 & 0.8637 \\
t-SNE  & 0.9290 & 0.9296 & \textbf{0.9098} & 0.9010 \\\hline

NDCM  & 0.9223 & 0.9225  & 0.5304 & 0.5309\\
DCM  & 0.9385 & 0.9434  & 0.8481 & 0.8941\\
\textbf{MFM} &  \textbf{0.9589} & \textbf{0.9645} & 0.8764 & \textbf{0.9117} \\\hline
\end{tabular}
\label{tab:proj_tab}
\end{table}

\subsection{Evaluation}

\begin{figure}[t]
\centering
\includegraphics[width=0.485\textwidth]{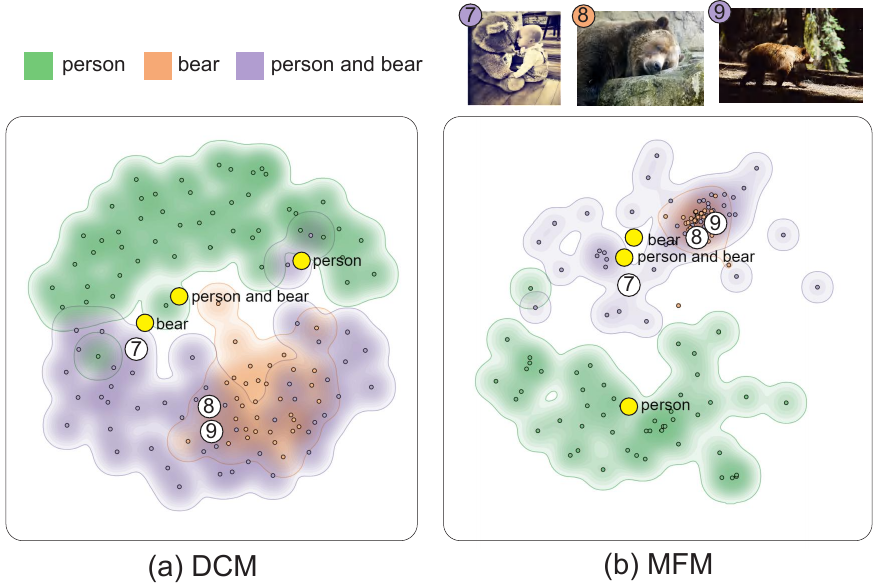}
\vspace{-5mm}
\caption{
DCM insufficiently captures the rank order of cross-modal distances between text and image embeddings, resulting in rather even distribution of image embedding points around the concept text embeddings, making it harder to observe differences in distribution pattern.
}\vspace{-3mm}
\label{fig:projection_compare2}
\end{figure}

\noindent
\textbf{Qualitative Comparison}.
As shown in Fig.~\ref{fig:projection_compare1} and Fig.~\ref{fig:projection_compare2}, MFM has many advantages for displaying both intra-modality and inter-modality features compared to the DCM method and traditional projection methods like MDS and t-SNE.
Specifically, Fig.~\ref{fig:projection_compare1} displays an intra-modal case with the projection of CLIP image embeddings for samples of 6 classes in CIFAR-10 dataset. 
The colors represent the zero-shot classification results based on CLIP.
Among the results, we can see that t-SNE achieves the best separation effect. 
However, t-SNE also has significant drawbacks in understanding the embeddings and identifying misalignment because it does not consider cross-modal features.
First, t-SNE is weaker at showing contextual information, such as the relation between different sets.
For example, we can find in Fig.~\ref{fig:projection_compare1}, the frog set (green point 1) can be confused with the bird set (yellow point 2) because of similar color or background, yet the t-SNE projection does not clearly show the relation compared to MFM.
In addition, our joint projection also shows better within-set distribution than t-SNE.
For example, with MFM, we can clearly see outliers or border points within sets (\eg blue point 5 and green point 6).
Point 5 corresponds to an image of a car driving on a highway, while most other car images are static scenes of parked cars.
Point 6 is a long-necked ostrich that is quite different in appearance from other birds.
However, these points are hard to identify in the t-SNE projection.
The MDS result in Fig.~\ref{fig:projection_compare1} (b) is more effective than t-SNE for showing the pointwise relationship, but the clustering effect is apparently weaker than t-SNE and MFM.
In addition, MDS tends to distribute the points quite evenly in the projected space, which compromises the display of in-set distribution and outliers. 
The DCM method shows the contextual set relationship better than t-SNE and displays set outliers only slightly better than MDS, since for the image modality, both DCM and MDS use metric loss, but MFM achieves better effect in both aspects.
Additionally, we compute Z-Score for a data point to help verify whether a visual outlier or border point is indeed so in the original high-dimensional space (see supplementary for detail).
We find that point 6, the most obvious outlier in Fig.~\ref{fig:projection_compare1} (d) indeed has the highest Z-score of 1.2379.

Fig.~\ref{fig:projection_compare2} shows the inter-modal case with the projection of CLIP text embeddings of three text queries together with the image embeddings of the query results.
Regarding the DCM results (Fig.~\ref{fig:projection_compare2} (a)), it does successfully merge the modality.
However, as shown in both cases, DCM has a significant weakness in that it scatters out the image embedding points quite evenly across the space, making it difficult for users to find distributional differences between different regions.
In comparison, for example, in Fig.~\ref{fig:projection_compare2} (b), we can find an obvious dense cluster in MFM results which contain many similar images of bear in the wild, but this pattern is less evident in DCM results.
In addition, MFM also more clearly shows the relation between concepts than DCM, as we can find that the query results of \emph{bear} and \emph{person and bear} have obvious overlap in both DCM and MFM results, signifying the closer relationship between these two text concepts, but the relative position of text embeddings in MFM is more coherent to this relation.

\noindent
\textbf{Quantitative Evaluation}.
To show some quantitative evidence of the advantage of MFM, we evaluate the method on COCO dataset using the trustworthiness metric and continuity metric~\cite{kaski2003trustworthiness, xia2022interactive}, where the former calculates how faithfully the projected kNN reflects the true kNN's in the embedding space and the latter calculates how well the original high-dimensional kNN is preserved in the projection.
Particularly, we calculate both inter-modal and intra-modal kNN for k=30.
However, in our scenario the inter-modal metric is more important as it measures the methods' ability to preserve multi-modal embedding structure.
For the evaluation process, we perform multiple rounds ($r=500$) of evaluation where in each round we randomly sample 500 images from COCO and project them together with the 80 category text embeddings in COCO object labels.
The final metric is the average of the results in all rounds.
As shown in Table~\ref{tab:proj_tab}, the experimental results indicate that MFM method performs consistently better than all the other methods in inter-modal truthworthiness and continuity, with higher than 2\% margin over the strongest baseline DCM.
In addition, MFM also achieves good performance in intra-modal metrics, only second to t-SNE.
NDCM~\cite{choo2012heterogeneous} is another fusion method using fully nonmetric objective.
We can see that among the three fusion methods (MFM, DCM and NDCM), our MFM is consistently better across inter-modal and intra-modal metrics, while fully nonmetric fusion method has significant disadvantage in keeping the intra-modal features.
We also need to note that the inter-modal metrics for non-fusion traditional methods like t-SNE and MDS cannot fully reflect their weakness in inter-modal scenarios because the modality gap will cause large distances between image embeddings and text embeddings in the projection space, making it difficult to perceive the differences between the inter-modal distances\cite{liang2022mind, zhou2023clip, ouali2023black}.

\section{\tool System}

\subsection{Visualization Interface}

\begin{figure}[t]
\centering
\includegraphics[width=0.485\textwidth]{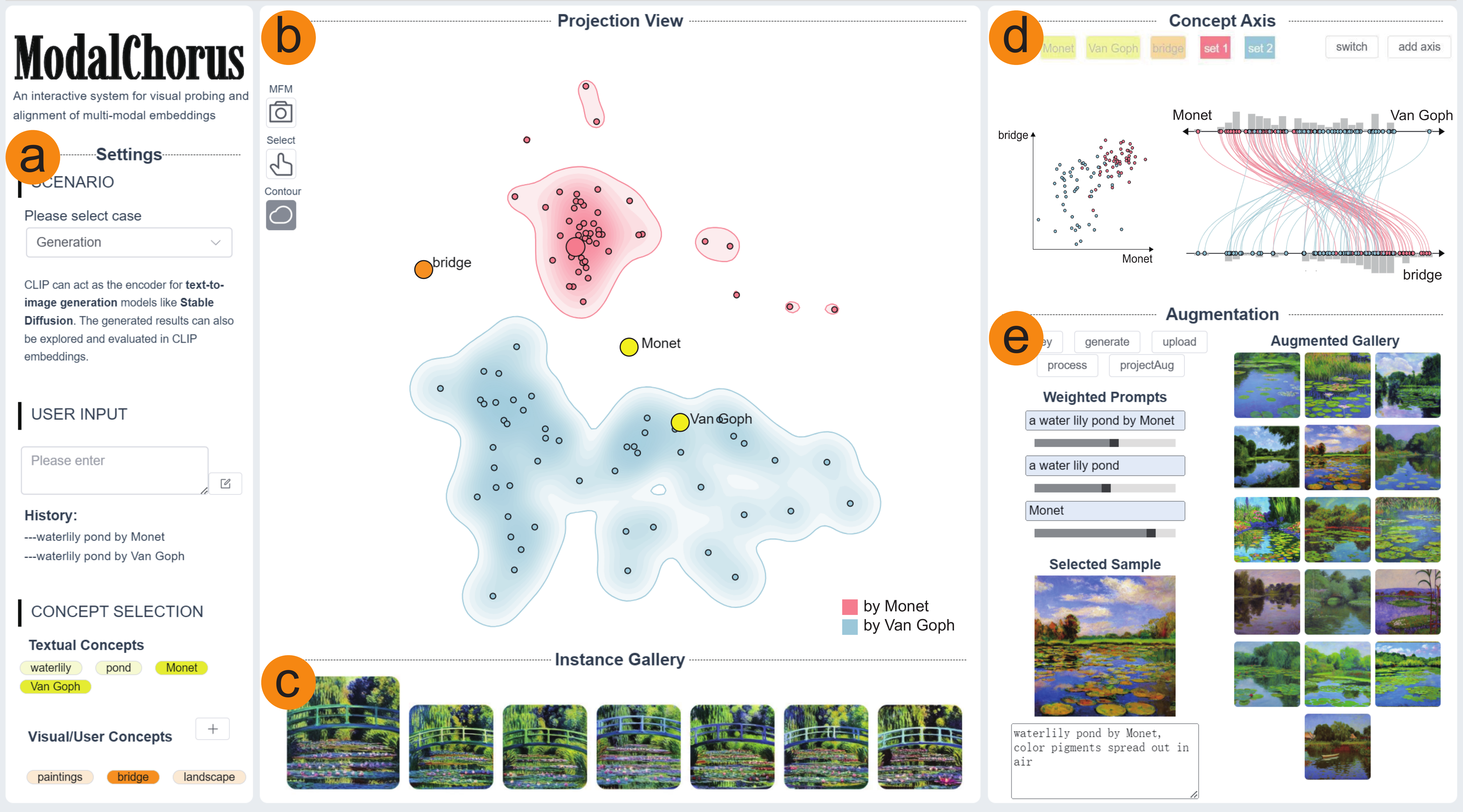}
\vspace{-6mm}
\caption{
ModalChorus system. (a) Settings panel on the left allow users' choice of task and dataset. 
The main projection view (b) displays the MFM dimension reduction result of embeddings.
The concept axis view (d) supports axis-based exploration, while the augmentation panel (e) facilitates uploading, generating, and tagging additional data for alignment. 
}
\label{fig:interface}
\end{figure}

\noindent
\textbf{Settings Panel}.
The settings panel (Fig.~\ref{fig:interface} (a)) allows users to specify some basic settings for their exploration, including tasks and inputs. 
Users can also select specific concepts in their input to produce contextual visualization in the projection view. 
Instead of relying solely on textual concepts explicitly extracted from existing text labels or prompts, ModalChorus extracts implicit concepts from images to provide a comprehensive display of concepts. To achieve this, we first leverage BLIP-2~\cite{li2023blip}, a multi-modal language model capable of receiving images as input and generating textual descriptions of those images. We then employ the TopicRank~\cite{bougouin2013topicrank} algorithms to extract candidate visual concepts based on the text generated by BLIP-2.

\noindent
\textbf{Projection View}.
The projection view (Fig.~\ref{fig:interface} (b)) is the main view of the system leveraging our proposed Modal Fusion Map to help users probe the embedding with different tasks and data.
Users can choose to turn on or turn off the contour to emphasize set relation or facilitate instance exploration respectively.
The projection view also includes an instance retrieval subview below (Fig.~\ref{fig:interface} (c)).
Users can mouse over the embedding point to see the corresponding image in the gallery.
They can also click the point to retrieve similar images to the selected one.
In addition, users can select a subset of points by lasso or ctrl-click, as shown in Case 2 and Fig.~\ref{fig:retrieval} in Sect.~\ref{sec:case}.

\begin{figure}[t]
\centering
\includegraphics[width=0.48\textwidth]{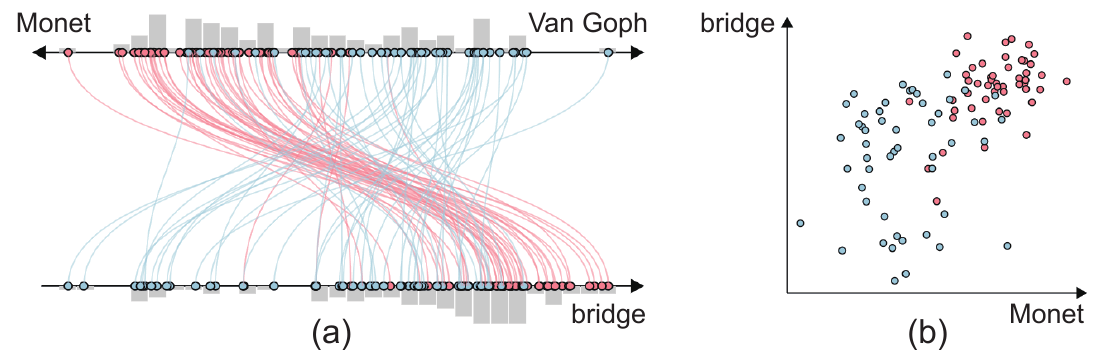}
\vspace{-2mm}
\caption{
The concept axis view allows the definition of unidirectional and bidirectional concept axes, showing the distribution of image embedding points in relation to concepts by similarity to concept text embedding.
Curves linking data points on different axes display the correlation between the distribution and reveal patterns such as entanglement.
}\vspace{-3mm}
\label{fig:concept}
\end{figure}

\noindent
\textbf{Concept Axis View}. 
As shown in Fig.~\ref{fig:interface} (d) and Fig.~\ref{fig:concept} (a), the concept axis view supports axis-based exploration of image embeddings in relation to text embeddings for user-selected concepts from the settings panel.
Users can define one-end axis with a single concept (\eg, bridge) or two-end axis with opposing concepts they want to contrast (\eg, Monet and Van Goph).
For one-end axis, the position of an image embedding point $x$ is:
\begin{gather}
\mu_{A}(x)=l\cdot\frac{sim(x, A)-min(sim(\hat{x}, A))}{max(sim(\hat{x}, A))-min(sim(\hat{x}, A))},
\end{gather}
where $sim(x, A)$ denotes the cosine similarity between $x$ and text embedding of concept $A$ in embedding space, $l$ is the length of the axis. 
For two-end axis, the position of $x$ is calculated as $l\cdot(0.5+\frac{\mu_{A}(x)-\mu_{B}(x)}{\mu_{A}(x)+\mu_{B}(x)})$.
When users define more than one axis, we use curves connecting the same instance on two axes to show the correlation.
Histogram is also used to help users see the overall distribution.
Apart from displaying instances of image embeddings, the concept axis can also represent the whole set or subset as small box at the average position of all the in-set points, showing users the mean value of the set and supporting further set-based alignment interaction as described in Fig.~\ref{fig:interact} and Sect.~\ref{ssec:align}.
We also allow users to switch to a scatterplot visualization (Fig.~\ref{fig:concept} (b)).

\noindent
\textbf{Augmentation Panel}.
The data augmentation panel (Fig.~\ref{fig:interface} (e)) supports interactive augmentation of alignment data.
In some alignment scenarios, users cannot find proper alignment data from the original dataset (for example, users may not find any satisfactory results generated by a pre-trained generative model).
For such a problem, the augmentation panel first allows users to upload a subset of samples to supplement the alignment data.
For unlabeled raw image data uploaded, this panel also integrates an auto-tagging function based on CLIP-interrogator~\cite{clip_inter}, which can generate tags associated with the image to enhance the alignment performance. 
Second, in cases where users even find it difficult to collect their own data, the augmentation panel also incorporates a generation function that enables users to leverage the weighted sum of existing text embeddings~\cite{chung2023promptpaint, wu2023uncovering} in a pre-trained generation model to synthesize more candidates of intention-aligned samples.

\subsection{Interactive Alignment}\label{ssec:align}
\begin{figure}[t]
\centering
\includegraphics[width=0.485\textwidth]{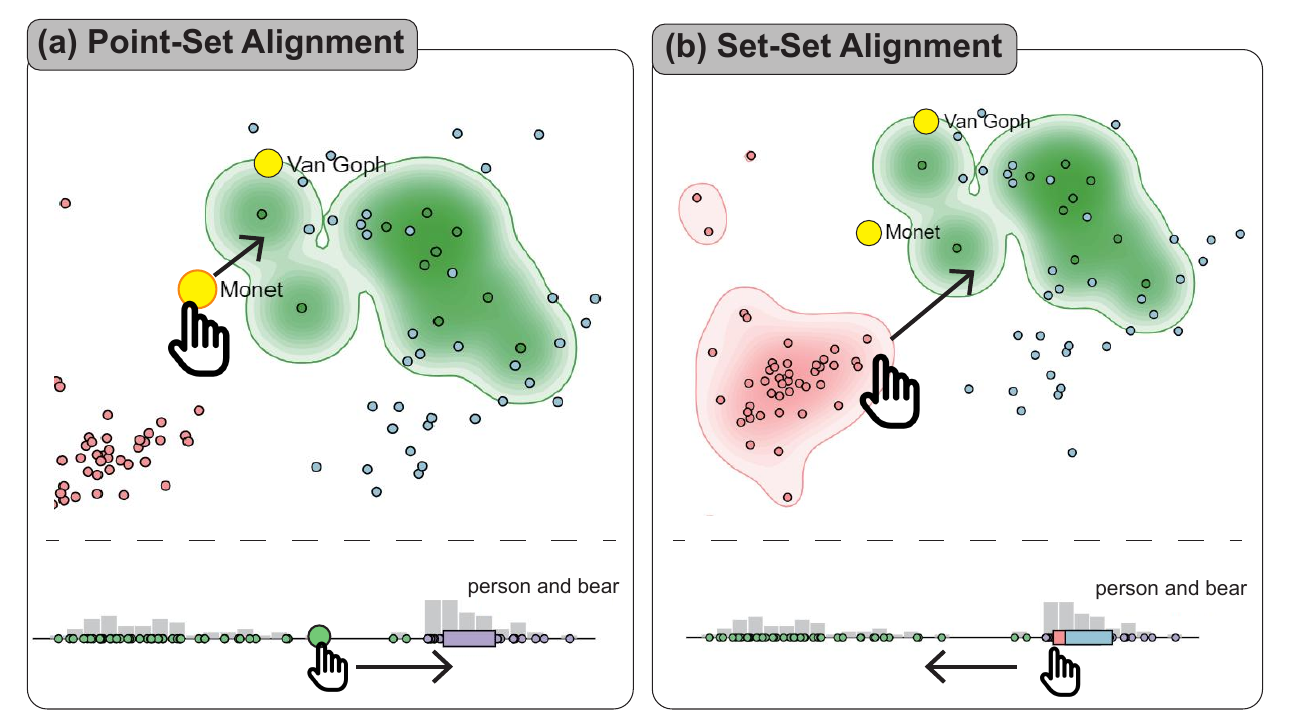}
\vspace{-5mm}
\caption{
We design interactions that allow users to visually express their alignment intentions, including point-set alignment and set-set alignment 
performed in both projection view and concept axis view.}
\label{fig:interact}
\end{figure}

\noindent
\textbf{Alignment Interaction Design}. 
We design a series of visual alignment interactions, which allow users to intuitively express diverse alignment intentions through visual metaphor and trigger backend fine-tuning without writing complex training code.
As shown in Fig.~\ref{fig:interact}, our interaction scheme supports set and point level intentions for alignment.
First, the common types of alignment mainly concern data points or subsets of points, which we categorize into two types: point-set alignment and set-set alignment, as shown in Fig.~\ref{fig:interact}.
First,
\emph{point-set} alignment encompasses various scenarios of aligning a set with a data point and vice versa.
For example, when users want to align a subset of retrieval results with a query text embedding or when users want to align a prompt embedding to fine-tune samples provided by themselves.
As shown in Fig.~\ref{fig:interact} (a), point-set alignment can be performed on either the projection view or the concept axis.
Formally, the high-level idea of point-set alignment can be summarized as follows:
Suppose we have a CLIP-based model $F(\cdot)$ which can map input text or image to different sets $C_{1}, C_{2}, ..., C_{N}$ in the sampled data.
For example, in classification, $C_{i}$ corresponds to the set of embeddings for a predicted class; in retrieval and generation, $C_{i}$ corresponds to the set of embeddings for the results of a single query or prompt. 
Given a user-selected misaligned image or text point $p$,
the target of point-set alignment is to tune the weights of $F(\cdot)$ such that $\hat{F}(p)$ is closer to the correct set $C_{i}$ in the embedding space.
Although the concrete implementation may vary for different tasks, in terms of the merged distance matrix, the effect is equivalent to achieving the following contrastive objective:
\begin{gather}
\frac{1}{|C_{i}|}\sum_{v \in C_{i}} M_{\hat{F}(p),v}<  \frac{1}{|C_{j}|}\sum_{u \in C_{j}} M_{\hat{F}(p),u} \ \ \forall j \neq i,
\end{gather}
where the estimated distance between $F(p)$ and $C_{i}$ should be smaller than any other set $C_{j}$.
Second, \emph{set-set} alignment involves moving two subsets of points closer or further in the concept axis or projection view.
Such alignment is intended to close the gap between two sets or distributions in the embedding space, or contrast two sets for distinguishing them better, which can be useful for cases like merging or disentangling concepts in retrieval or generation.
As shown in Fig.~\ref{fig:interact} (b), in the projection view, they can drag a set contour towards another, while in the axis view, they can drag one set box closer to or away from another.
Formally, the high-level idea of set-set alignment is:
Suppose users identify a misaligned set or subset of embeddings $C_{e}$, 
where $C_{e}$ is not align with the input $p$.
Next, users find another correct set $C_{i}$ either by visual exploration of other projected data points or by data augmentation.
The goal of set-set alignment can then be formulated as:
\begin{gather}
\frac{1}{|\hat{F}(p)|\cdot|C_{i}|}\sum_{v \in \hat{F}(p), u \in C_{i}}M_{u,v}<\frac{1}{|\hat{F}(p)|\cdot|C_{e}|}\sum_{v \in \hat{F}(p), u \in C_{e}}M_{u,v}.
\end{gather}

\noindent
\textbf{Alignment Fine-tuning Implementation}.
Our system provides a general framework to map users' visual interactions shown in Fig.~\ref{fig:interact} to backend fine-tuning operations that align the model's output in the embedding space. 
As the visual representations are decoupled from actual backend implementation, our framework can incorporate any kind of specific fine-tuning methods.
For demonstration purposes, our study implements two methods.
First, for the classification and retrieval cases, we implement triplet loss~\cite{schroff2015facenet} based alignment.
Second, for the generation cases, we implement the low-rank adaptation method~\cite{hu2021lora}.
More detail is provided in the supplementary material.

\section{Case Studies}\label{sec:case}
In this section, we perform three case studies to demonstrate the usefulness of the Modal Fusion Map and ModalChorus system, which cover three different tasks based on multi-modal embeddings, including zero-shot classification, text-to-image retrieval, and generation.
Particularly, we demonstrate how our visual probing integrates with and enhances interactive few-shot alignment~\cite{gao2024clip, huang2024lp++}.

\subsection{Case 1: Zero-shot classification}
In this case, we demonstrate how our system can be used to visualize the zero-shot classification~\cite{radford2021learning} based on multi-modal embedding clustering and help refine the embedding interactively with one-shot point-set alignment.
Specifically, we use the CIFAR-10 image classification dataset~\cite{krizhevsky2009learning} to show an example.
Here we suppose no ground-truth labels are available.
This is to simulate real-time analysis of zero-shot embedding-based classification in the wild for unknown data, where interactive visual analysis with human intervention is most helpful.

\begin{figure}[t]
\centering
\includegraphics[width=0.485\textwidth]{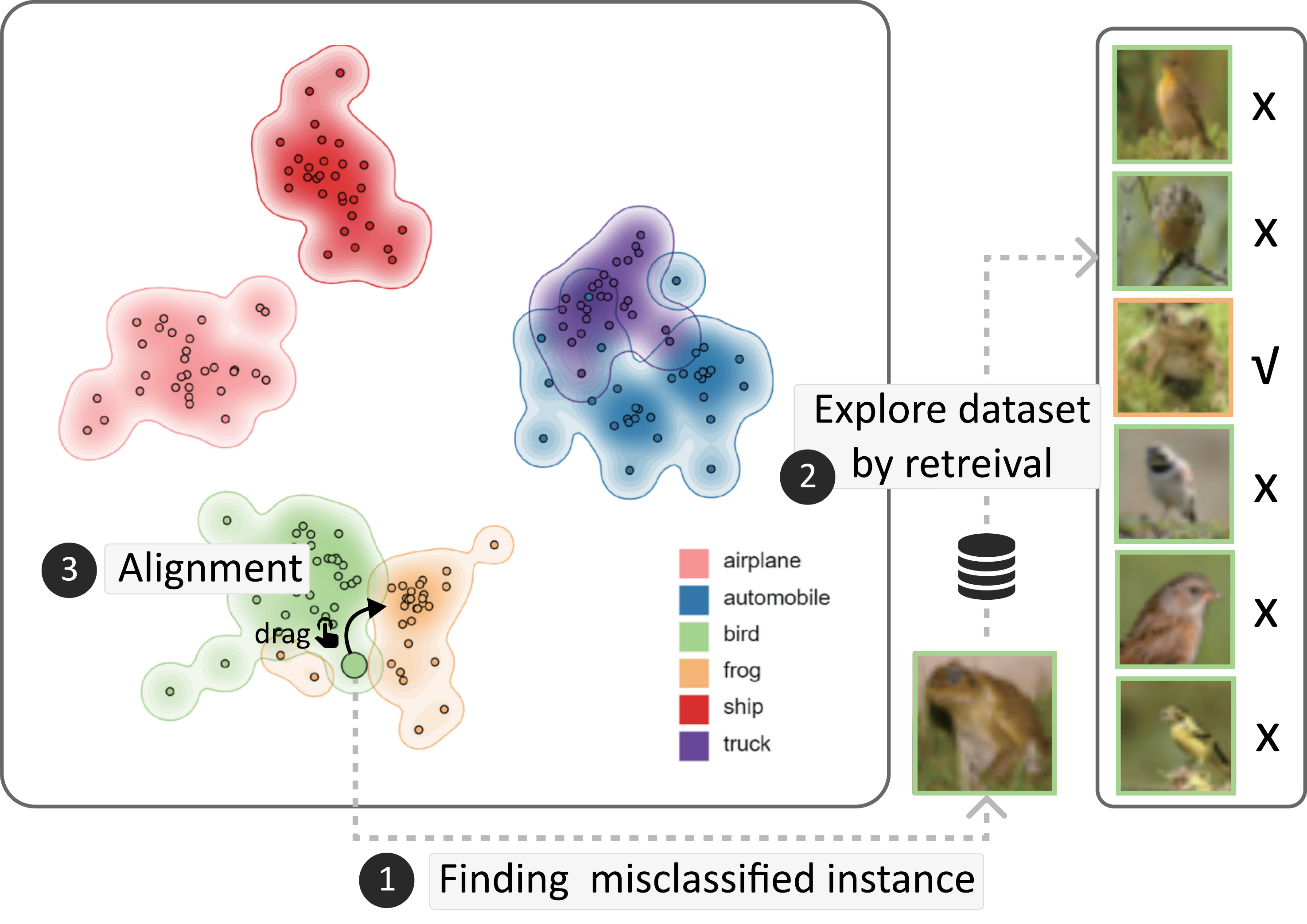}
\vspace{-6mm}
\caption{
In zero-shot classification with CLIP, users can leverage \tool to identify potential examples of misalignment (in this case, classification mistake) and perform point-set alignment to refine CLIP.
}
\label{fig:classification}
\end{figure}

Users first select the classification task and the dataset.
Then, users subjectively select some classes that they suspect may be confusing for CLIP, including classes of small wild animals and classes of vehicles.
Specifically, they select 6 class concepts they want to explore, including \emph{airplane}, \emph{automobile}, \emph{truck}, \emph{ship}, \emph{bird} and \emph{frog}.
Then, the system predicts the class of each image according to the cross-modal proximity between the image embedding and class text embedding.
The system samples the data for visual probing.
Specifically, it retrieves 50 closest images to each class text based on the CLIP embeddings.
They can see that the CLIP embedding sets of the \emph{bird} and \emph{frog} clusters are indeed quite close, and they can find that the highlighted point corresponding to the data item selected in the concept view is at the border of the \emph{bird} cluster.
In fact, when users see the image, they find that this item is actually \emph{frog} but misclassified as \emph{bird}.
By retrieving similar images in the instance retrieval view, users can further understand that this is because there are some birds and frogs with similar colors and outlines.
Subsequently, users perform point-set alignment by dragging this point closer to the correct \emph{frog} cluster, as shown in Fig.~\ref{fig:classification}.

In the backend, we use the ground truth to verify that our visual alignment is indeed helpful.
Specifically, before the alignment, the overall accuracy is $69.28\%$.
Particularly, the category with the lowest accuracy is \emph{frog}, with only $32.82\%$.
After the visual alignment of only one single data point in this case, the accuracy for \emph{frog} category rises to $45.24\%$ among the 10,000 images in batch 1 of CIFAR-10. At the same time, overall accuracy also increases to $70.66\%$.

\subsection{Case 2: Instance Retrieval and Compositional Logic}
In this case, we show how our system can be used to visualize and refine the compositional logic in instance retrieval. 
Specifically, we test cross-modal retrieval on the COCO 2017 dataset~\cite{lin2014microsoft}. 
The training set of the COCO dataset consists of more than 110,000 images with annotations of captions and objects in the image from 80 categories.

\begin{figure}[t]
\centering
\includegraphics[width=0.485\textwidth]{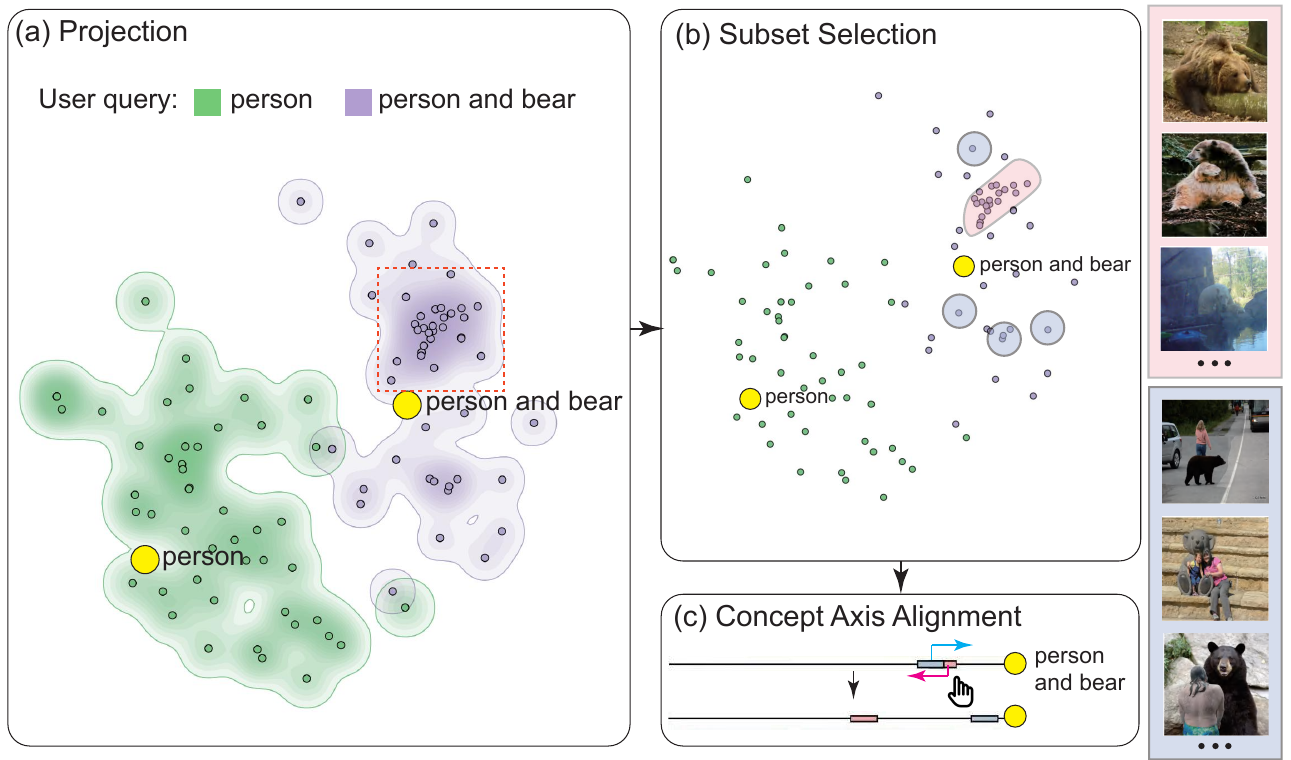}
\vspace{-6mm}
\caption{
In cross-modal retrieval with CLIP, users can leverage \tool to visualize both the query text and image results embeddings.
Users can further select subsets of data and perform axis-based alignment.
}
\label{fig:retrieval}
\end{figure}

For the retrieval, users enter two queries: the first one is a simple keyword: \emph{"person"} while the second one is composed of two elements using natural language logic expression: \emph{"person and bear"}.
Users first leverage our MFM method to project the respective top 50 image results of the two queries in relation to the keyword texts in CLIP embeddings.
In the contextual projection shown in Fig.~\ref{fig:retrieval} (a), users can see that there is an obvious dense cluster near the text embedding of the composed query \emph{"person and bear"}.
In contrast, the distribution of image embeddings is rather sparse near the text embedding of the simple query \emph{"person"}.
When users mouse over some data points to explore particular instances, they find that the retrieval results of \emph{"person"} are more diverse than those of \emph{"person and bear"}.
More importantly, they even find that the cluster actually contains many similar incorrect results only containing bear in the wild without any person, which shows that the CLIP embeddings do not sufficiently understand the logic in \emph{"person and bear"}.
Users can further verify this finding by adding another keyword query \emph{"bear"} and visualize the results together, as shown in Fig.~\ref{fig:projection_compare2} (b).

Upon identifying the misalignment issue for the composed query, users can proceed to interactively align the CLIP embeddings in the system's align mode.
Specifically, in the projection view, they first lasso to select samples of the incorrect cluster, which are added to the first alignment subset represented by pink color. 
Next, they also discover some individual samples of correct images containing both person and bear near the text embedding, which are added to the second alignment subset represented in blue.
Subsequently, users can see that the incorrect subset and correct samples are quite close and hard to separate.
Finally, users can perform set alignment by dragging the incorrect subset farther away, triggering fine-tuning in the backend.

After the alignment, users can exploit the new CLIP embeddings to re-rank the previous top 500 results.
We implement re-ranking instead of completely indexing all the data points in the dataset for faster system reaction. This only takes a few seconds, together with the few-shot alignment.
To verify quantitatively that such alignment is indeed helpful, in the backend, we calculate the top k accuracy of the new results compared to previous results, as shown in the table.

\begin{table}
\scriptsize
\centering
\caption{Accuracy of \emph{"person and bear"} before and after the alignment.}
\vspace{-2mm}
\begin{tabular}{|c|c|c|c|c|c|}
\hline
& top 5 &  top 20 & top 30& top 40 & top 50\\\hline
\textbf{Before} & 60.00\% & 50.00\%& 43.33\% & 40.00\% & 40.00\%\\\hline
\textbf{After} & \textbf{80.00\%} & \textbf{60.00\%}& \textbf{53.33\%}&\textbf{52.50\%}&\textbf{52.00\%}\\\hline
\end{tabular}

\label{tab:quant}
\end{table}

\begin{figure*}[t]
\centering
\includegraphics[width=0.995\textwidth]{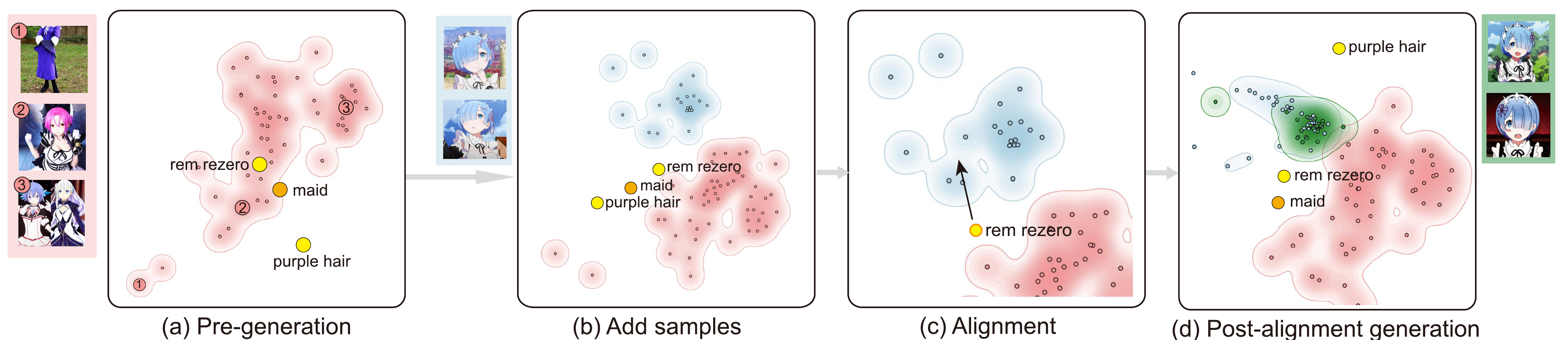}
\vspace{-2mm}
\caption{
Alignment for text-to-image cross-modal generation. 
The first case shows concept injection, where the original model does not understand a concept, and users provide visual samples to align with the textual concept keywords.
}
\label{fig:gen}
\end{figure*}

\subsection{Case 3: Concept Injection and Disentanglement in Cross-modal generation}
In this case, we show how our system can be used for alignment in cross-modal generation, with examples of aligning text-to-image Stable Diffusion model.
Particularly, compared to specialized AIGC fine-tuning tools such as IntentTuner~\cite{zeng2024intenttuner}, which only focuses on data augmentation and training functions like LoRA~\cite{hu2021lora} and DreamBooth~\cite{ruiz2023dreambooth}, our visual probing framework allows users to visually inspect and compare the generation results, augmentation data, and prompt keywords before and after fine-tuning through embedding visualization.
We choose Stable Diffusion V1-4, which uses CLIP as the input encoder.

The first example (Fig.~\ref{fig:gen}) shows the case of alignment for concept injection, where the pre-trained model does not understand a concept, and users try to inject it into the model's knowledge.
For example, as shown in Fig.~\ref{fig:gen}, users may want to input prompt \emph{"Rem rezero"}, which is the name of an animation character, to generate images of the character.
However, after the generation of 50 samples by the original model, 
users can find that our system detects some visual keywords such as \emph{"purple hair"} that is unexpected since the desired character has a prominent feature of short blue hair.
Users also enter another concept keyword \emph{"maid"}, which describes the signature dressing style of the expected character.
Then, MFM produces a joint projection of the keywords and the generated images, as shown in Fig.~\ref{fig:gen} (a).
Users can find in this projection view many abnormal results.
For example, the outliers like Fig.~\ref{fig:gen} (a) (1) are images of realistic photos.
Users can also confirm that some images close to the \emph{"purple hair"} text embeddings contain distinct purple hair from the expected blue hair (Fig.~\ref{fig:gen} (a) (2)).

To align the model with this new concept, users collect 20 correct sample images of Rem and upload them to the system.
The newly uploaded samples are added to the projection as shown in Fig.~\ref{fig:gen} (b).
We can see the projection view clearly shows the gap between the pre-generated results (red) and the correct samples (blue), while the prompt keyword \emph{"rem rezero"} sits somewhere in the middle between the two clusters, indicating insufficient alignment to the correct set.
After observing this, users can directly drag the text embedding point of \emph{"rem rezero"} towards the correct sample set, as shown in Fig.~\ref{fig:gen} (c).
This action triggers the backend alignment process.
When the alignment is finished, the system generates 50 samples (green) of \emph{"rem rezero"} with the newly-aligned model (Fig.~\ref{fig:gen} (d)).
We can see that compared to the red set, the green set are much closer to the correct sample cluster, which signifies successful alignment to the correct concept.
Users further explore the details of the new generated images and find that they have captured the most important features of the desired character, including the signature short blue hair and the maid dress.

The second example (Fig.~\ref{fig:teaser}) showcases alignment for disentangling the generation.
Sometimes, misalignment in the cross-modal generation model causes concepts to entangle, leading to unexpected and uncontrollable generation.
For instance, when users want to generate landscape paintings, they enter two prompts with the same subject but different artists' names: \emph{"waterlily pond by Van Goph"} and \emph{"waterlily pond by Monet"}.
The system first generates 50 samples for each prompt with the same set of random seeds.
Apart from textual keywords like \emph{"Monet"} and \emph{"Van Goph"}, the system also detects an unexpected visual concept: \emph{"bridge"}.
Then, by exploring our MFM projection view as shown in Fig.~\ref{fig:teaser} (a), users can find that the points generated by \emph{"waterlily pond by Monet"} (red) are concentrated in a dense cluster close to the text embedding of \emph{"bridge"} while the results of \emph{"waterlily pond by Van Goph"} are distributed more sparsely in the projection space.
Inspecting the data points, users can discover that the Monet results have a highly similar composition, almost always containing a bridge.
This pattern is more evidently shown in the concept axis in Fig.~\ref{fig:teaser} (a), where the high values on the Monet dimension are strongly bundled with high values on the bridge dimension for the red set.
In contrast, the Van Goph results are much more diverse with different compositions.
This observation indicates that in the Monet prompt, the name of the artist is highly entangled with the visual concept of the bridge, thus significantly reducing the diversity of the generation.
To align the model for disentanglement,
users first need some fine-tuning data but may find it difficult to collect Monet's paintings of the waterlily pond manually.
To address the issue, they can exploit the weighted embedding function provided by our augmentation panel to generate more disentangled samples.
Specifically, users can combine the CLIP embeddings of different keywords and phrases in the original prompt through a weighted sum of the embedding vectors to guide the generation of augmented samples.
Users can select from these generated augmentation images satisfactory samples that match the prompt \emph{"waterlily pond by Monet"}.
They can then add the samples to the projection (green set) in Fig.~\ref{fig:teaser} (b), where users can find an obvious distance between the pre-generated Monet set and the augmentation set.
Next, as shown in Fig.~\ref{fig:teaser} (c), users can drag the pre-generated Monet set contour towards the augmentation set in the projection or drag the box representing the pre-generated Monet set in the concept axis, which triggers the backend alignment process of the two sets.
Finally, in Fig.~\ref{fig:teaser} (d), the system will generate a new set (purple) by the same prompt of the original red set (\emph{"waterlily pond by Monet"}) with the post-alignment model.
Users can see that compared with the red set, the purple set is more aligned to the green set while having more diversity (containing images with and without bridges).

\section{Discussion}

\noindent
\textbf{Speed limitations}.
Our system and method have two limitations in terms of speed.
First, even though the parametric method can scale to large datasets with shorter asymptotic time (as shown in supplementary), for smaller datasets, it is not as fast as some traditional methods like t-SNE.
Second, the speed of the alignment fine-tuning depends on the specific implementation for different tasks.
For classification and retrieval tasks, the triplet loss-based fine-tuning only takes a few seconds.
However, for the generation task, the commonly used LoRA fine-tuning can take a few minutes.
To address this, we can take advantage of the latest accelerated fine-tuning methods such as HyperDreamBooth~\cite{ruiz2024hyperdreambooth}.

\noindent
\textbf{Scalability to more modalities}. 
In this study, we only test our system and method on embeddings of two modalities.
However, some multi-modal embeddings involve more than two modalities.
For example, the ImageBind~\cite{girdhar2023imagebind} embedding models incorporate six modalities including images, text, audio, depth, thermal, and IMU data.
For these data, we can extract multi-modal semantic features as concepts and extend our concept visualization to cover more modalities.
For modalities that are difficult to observe visually, such as audio, we can represent their conceptual features using text or images.
Our MFM method can also be improved to visualize different pairs of modalities, such as text-audio and image-audio.

\noindent
\textbf{Pixel-level alignment}.
Even though our study enables various set and point level alignments, sometimes these alignments are not enough for fine-grained cross-modal tasks.
For example, embedding-based objection detection requires sub-instance pixel level alignment~\cite{zhong2022regionclip}.
Our research has not so far touched upon this type of sub-instance alignment.
Regarding this problem, in future work, our system can integrate interactive semantic segmentation such as Segment Anything Model~\cite{kirillov2023segment} into the alignment process to allow users to emphasize certain parts of the image they want the model to understand.

\section{Conclusion}
In this study, we propose a visual probing and alignment framework for exploring and interactively refining multi-modal embeddings.
Particularly, for visual probing, we address the modality gap problem by developing a dimension reduction method called Modal Fusion Map (MFM) to optimize the display of inter-modal embedding features.
To facilitate interactive alignment, we design an interaction scheme supporting various alignment intentions including point-set and set-set alignment.
As shown in our quantitative evaluation and case studies, our framework can help intuitive visual probing and alignment for diverse tasks.
This shows the opportunities for future research to increase human moderation of large models that are growing in size but decreasing in transparency.

\acknowledgments{%
The authors wish to thank the anonymous reviewers for their valuable comments. This work is supported partially by National Natural Science Foundation of China (62172398), and the Guangzhou Basic and Applied Basic Research Foundation (2024A04J6462, 2023A03J0142).
}

\bibliographystyle{abbrv-doi-hyperref}

\bibliography{template}








\end{document}